\documentclass[nohyperref]{article}

\usepackage{microtype}
\usepackage{graphicx}
\usepackage{subfigure}
\usepackage{booktabs} 
\usepackage{multirow}
\usepackage{hyperref}
\usepackage{makecell}

\usepackage[accepted]{icml2022}

\usepackage{amsmath}
\usepackage{amssymb}
\usepackage{mathtools}
\usepackage{nicefrac}
\usepackage{amsthm}
\usepackage{bbm}
\usepackage[capitalize,noabbrev]{cleveref}

\theoremstyle{plain}

\theoremstyle{definition}

\theoremstyle{remark}

\usepackage[textsize=tiny]{todonotes}

\icmltitlerunning{ME-GAN for Multi-view ECG synthesis Conditioned on Heart Diseases}

\begin{document}

\twocolumn[
\icmltitle{ME-GAN: Learning Panoptic Electrocardio Representations for Multi-view ECG Synthesis Conditioned on Heart Diseases}



\icmlsetsymbol{equal}{*}

\begin{icmlauthorlist}
\icmlauthor{Jintai Chen}{equal,zjucs}
\icmlauthor{Kuanlun Liao}{equal,zjucs}
\icmlauthor{Kun Wei}{xidian}
\icmlauthor{Haochao Ying}{zjugw,kl}
\icmlauthor{Danny Z. Chen}{nd}
\icmlauthor{Jian Wu}{yiyuan}
\end{icmlauthorlist}

\icmlaffiliation{zjucs}{College of Computer Science and Technology, Zhejiang University, Hangzhou 310058, China;}
\icmlaffiliation{xidian}{School of Electronic Engineering, Xidian University, Xi’an 710071, China;}
\icmlaffiliation{zjugw}{School of Public Health, Zhejiang University, Hangzhou, China;}
\icmlaffiliation{kl}{The Key Laboratory of Intelligent Preventive Medicine of Zhejiang Province, Hangzhou, Zhejiang 310058, China;}
\icmlaffiliation{nd}{Department of Computer Science and Engineering, University of Notre Dame, Notre Dame, IN 46556, USA}
\icmlaffiliation{yiyuan}{The Second Affiliated Hospital School of Medicine, School of Public Health, and Institute of Wenzhou, Zhejiang University, Hangzhou 310058, China}
\icmlcorrespondingauthor{Haochao Ying}{haochaoying@zju.edu.cn}

\icmlkeywords{GAN, ECG synthesis}

\vskip 0.3in
]



\printAffiliationsAndNotice{\icmlEqualContribution} 

\begin{abstract}
Electrocardiogram (ECG) is a widely used non-invasive diagnostic tool for heart diseases. Many studies have devised ECG analysis models (e.g., classifiers) to assist diagnosis. As an upstream task, researches have built generative models to synthesize ECG data, which are beneficial to providing training samples, privacy protection, and annotation reduction. However, previous generative methods for ECG often neither synthesized multi-view data, nor dealt with heart disease conditions. In this paper, we propose a novel disease-aware \textbf{g}enerative \textbf{a}dversarial \textbf{n}etwork for \textbf{m}ulti-view \textbf{E}CG synthesis called ME-GAN, which attains panoptic electrocardio representations conditioned on heart diseases and projects the representations onto multiple standard views to yield ECG signals. Since ECG manifestations of heart diseases are often localized in specific waveforms, we propose a new \textit{mixup normalization} to inject disease information precisely into suitable locations. In addition, we propose a \textit{view discriminator} to revert disordered ECG views into a pre-determined order, supervising the generator to obtain ECG representing correct view characteristics. Besides, a new metric, \textit{rFID}, is presented to assess the quality of the synthesized ECG signals. Comprehensive experiments verify that our ME-GAN performs well on multi-view ECG signal synthesis with trusty morbid manifestations.
\end{abstract}

\section{Introduction}
Heart diseases are major threats to global health~\cite{abubakar2015global,roth2018global}.  Electrocardiogram (ECG), a non-invasive diagnostic tool for heart diseases, is widely-used in clinical practice~\cite{holst1999confident}, putting a heavy burden on cardiologists. To alleviate this stress, many research efforts were put into developing ECG classifiers~\cite{bian2022identifying,golany202112,kiranyaz2015real} for automatic ECG analysis, yielding considerable performances on limited tests. But, there were still concerns on developing automatic ECG classifiers, such as patient privacy protection~\cite{ecgadvgan,hazra2020synsiggan} and absence of well-annotated ECG signals~\cite{golany2020simgans}. Thus, a key upstream task~\cite{golany2020simgans,zhang2021synthesis} is to synthesize ECG signals in order to increase the diversity and quantity of training samples, which can also help reduce annotation needs if ECG signals are synthesized conditioned on heart diseases.

\begin{figure*}[t]
\vskip -.1 in
    \centering
    \includegraphics[width=\textwidth]{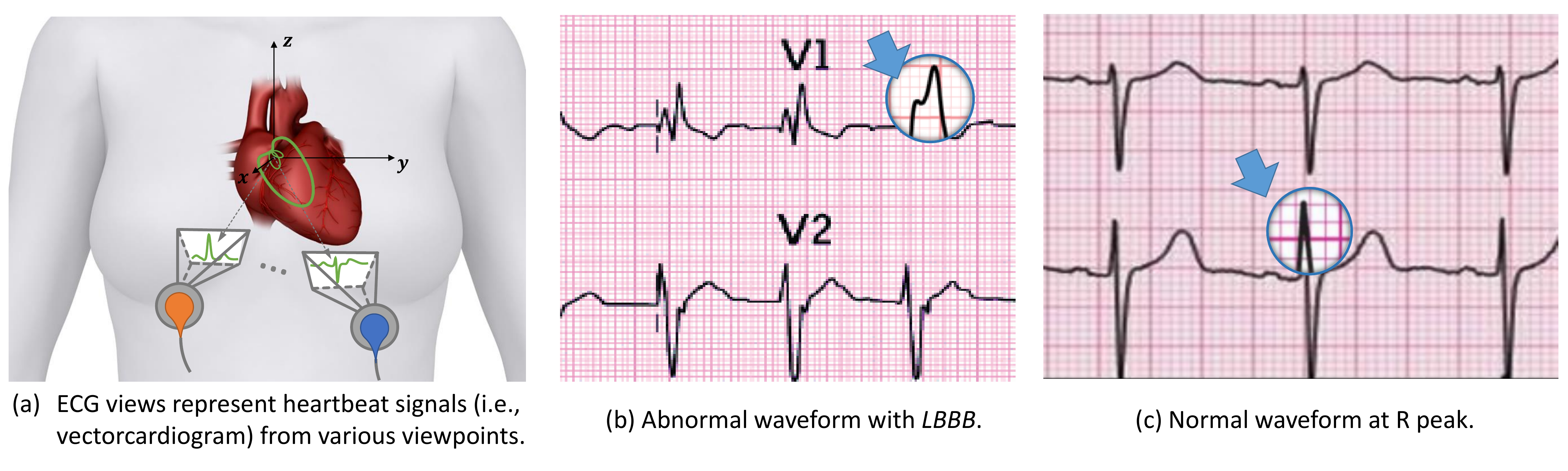}
    \vskip -.2 in
    \caption{(a) Illustrating the principle of multi-view ECG signal recording by projecting heartbeat signals along different viewpoints. A viewpoint is represented by two angles ($\theta, \phi$) (the polar and azimuthal angles; defined in Sec.~\ref{sec:ecgbackgrounds}). Comparing (b) and (c), the waveform changes caused by diseases can be localized (e.g., \textit{left bundle branch block abnormality} (\textit{LBBB}) can lead to notches on R waves).}
    \label{fig:ecg}
    \vskip -0.1 in
\end{figure*}

There are two main issues in the known ECG synthesis methods. \textbf{(1) Trusty multi-view ECG synthesis.} In clinical practice, ECG signals from different views are clinically useful~\cite{graybiel1946electrocardiography,case1979sequential}, representing heartbeat signals from various viewpoints (see Fig.~\ref{fig:ecg}(a)). However, previous ECG synthesis models often yielded single-view ECG~\cite{golany2019pgans,golany2020simgans,ecgadvgan,hazra2020synsiggan,zhu2019electrocardiogram} or independently synthesized different views without explicitly considering view correlations~\cite{kuznetsov2020electrocardiogram,mcsharry2003dynamical,thambawita2021deepfake}. Such synthesized ECG signals cannot provide trusty representations of heartbeats, resulting in limited applications.
\textbf{(2) ECG synthesis conditioned on heart diseases.} Unlike some global conditions used in image synthesis~\cite{stylegan} (e.g., the gender impacts synthesized face pictures globally), ECG manifestations of some heart diseases are often localized in specific waveforms~\cite{franklead}. For example, the \textit{left bundle block abnormality} (\textit{LBBB}) often presents notches on/near R waves (R wave is a typical waveform in ECG signals; see Fig.~\ref{fig:ecg}(b)).
But, most ECG generative methods either omitted disease information~\cite{hazra2020synsiggan,zhu2019electrocardiogram} or synthesized ECG with different heart diseases by developing different models~\cite{zhang2021synthesis,golany2020simgans}, which do not address the general heart disease embedding problem.

To address these two issues, this paper develops a novel \textbf{GAN} for \textbf{M}ulti-view \textbf{E}CG synthesis conditioned on heart diseases called ME-GAN, which consists of a generator, a major discriminator to distinguish real ECG and to predict the disease categories, and an auxiliary discriminator called \textit{view discriminator} to ensure the synthesized ECG representing correct view characteristics.

\textbf{To synthesize highly trusty multi-view ECG signals,} we devise the generator following the principle of ECG signal recording. 
As shown in Fig.~\ref{fig:ecg}(a), ECG signals in different views actually record electrocardio signals (representing heartbeats) from different viewpoints~\cite{franklead}. A study~\cite{chen2021electrocardio} has verified that a neural network can predict ECG views based on electrocardio representations with query viewpoints (represented by angles).
Motivated by 3D-aware GANs~\cite{zhou2021cips,tian2018cr}, our generator performs in two stages, which synthesizes new ``stereo'' electrocardio representations of heartbeat signals in stage one, and employs the Nef-Net decoder~\cite{chen2021electrocardio} to yield the standard ECG views based on the synthesized representations in stage two. To emphasize view correlations among the synthesized views, we randomly shuffle the view order and utilize a \textit{view discriminator} to revert the views into the pre-determined order. Thus, the \textit{view discriminator} supervises the generator to synthesize signals representing correct and differentiable view characteristics.

\textbf{To inject disease information,} we present a novel \textit{mixup normalization} layer in the generator to precisely control the degrees and locations of disease information embedding in the synthesized signals, since morbid manifestations are localized in specific waveforms.
Different from the self-modulation~\cite{chen2018self} and AdaIN~\cite{stylegan} that provided channel-wise normalization with global conditions, our \textit{mixup normalization} seeks the correct locations and performs length-wise normalization to inject disease information in the synthesized signals.

In addition, we present a pre-trained 1D Inception~\cite{szegedy2016rethinking} with a metric called \textit{relative Fr\'echet Inception Distance (rFID)} to assess the quality of synthesized multi-view ECG signals. The \textit{rFID} is analogous to the \textit{FID} metric for synthesized images~\cite{heusel2017gans}. 

The main contributions of this paper are as follows.

\textbf{(A)} We propose a novel GAN model called ME-GAN, which is the first model that is able to synthesize standard 12-view ECG signals and simultaneously considers heart diseases. Following the 3D-aware GAN strategy, ME-GAN first synthesizes the representations of heartbeat signals and then projects them into standard views to represent ECG signals. Our design follows the ECG recording principle and essentially ensures representations of ECG views to be consistent.

\textbf{(B)} We propose a novel \textit{mixup normalization} to precisely determine the locations and degrees to inject disease information into the synthesized ECG signals, based on the ECG characteristics of morbid manifestations.

\textbf{(C)} We present a novel \textit{view discriminator}, which reverts randomly shuffled ECG views into a pre-determined order, by processing heavily masked ECG signals. Thus, any locations on the synthesized ECG signals are ensured to represent correct view characteristics.

\textbf{(D)} We introduce a pre-trained 1D Inception with a new metric called \textit{rFID} to assess the quality of synthesized multi-view ECG signals. Comprehensive experiments verify that the synthesized ECG signals by our ME-GAN are highly trusty and are useful for improving ECG classifiers.

\section{Backgrounds}\label{sec:ecgbackgrounds}
\textbf{Electrocardiogram (ECG) and ECG GANs.} \ Clinical studies~\cite{graybiel1946electrocardiography,case1979sequential,grant1950spatial} indicated that single-view ECG signals cannot provide enough information for diagnosis and surgery, and ECG views (12 views as default) essentially represent the projections of heartbeat signals along various viewpoints (see Fig.~\ref{fig:ecg}(a)). But, most previous ECG synthesis methods~\cite{ecgadvgan,hazra2020synsiggan,zhu2019electrocardiogram,kuznetsov2020electrocardiogram,thambawita2021deepfake,kuznetsov2021interpretable} either yielded a single ECG view or independently synthesized multi-view ECG signals following various GAN models~\cite{wgan} while omitting view correlations. Besides, most of the known methods~\cite{zhu2019electrocardiogram,hazra2020synsiggan} ignored the presence of heart diseases, while only a few~\cite{zhang2021synthesis,golany2020simgans} synthesized ECG signals belonging to different diseases using different models.

\textbf{A Review of Electrocardio Panorama and Nef-Net.} \ It was empirically proved that multi-view ECG signals are projections of electrocardio signals along corresponding viewpoints~\cite{chen2021electrocardio}. Researchers often used a spherical coordinate system with the origin point at the central electric terminal of the heart, and the \textit{x}-axis, \textit{y}-axis, and \textit{z}-axis are defined by the anatomical sagittal axis, inverse frontal axis, and vertical axis, respectively. In the spherical coordinate system, the viewpoints of ECG views are represented by two angles, a polar angle $\theta$ and an azimuthal angle $\phi$. Based on this system, an auto-encoder, Nef-Net, was proposed~\cite{chen2021electrocardio} in which the encoder predicts the ``stereo'' electrocardio representations of a known ECG case and the decoder yields a new ECG view by projecting the representations along the query viewpoints. In this paper, we also utilize the spherical coordinate system (see Fig.~\ref{fig:ecg}(a)), represent viewpoints by $(\theta, \phi)$, and use the decoder as a part of our generator. Unlike Nef-Net which synthesizes new ECG views of known cases, our ME-GAN is developed to synthesize multi-view ECG of new cases. 

\textbf{Related Generative Adversarial Networks.} \ The generative adversarial network (GAN)~\cite{goodfellow2014generative} was first proposed for image synthesis, and was then applied in many fields~\cite{sutskever2014sequence,molgan}. Various studies~\cite{sutskever2014sequence,ramsundar2015massively} showed that the condition information contributed to synthesis effects. Typically, conditions were provided to generators by concatenation with random noises~\cite{cgan}. Later, additional supervisions were used in discriminators to promote the effects of condition embedding~\cite{acgan,chen2016infogan,odena2016semi}, which were widely used in medical data synthesis and obtained competitive performance~\cite{yi2019generative}. Recently, conditions were also introduced into GANs indirectly, by style transfer~\cite{huang2017arbitrary,stylegan2}, linear interpolation~\cite{radford2015unsupervised}, and latent vector modification~\cite{agrawal2021directional}.
However, these methods did not give strict definitions and bounds of the embedded ``conditions'', and were not suitable to biomedical fields. Other related work is on multi-view image synthesis. Following Nerf~\cite{mildenhall2020nerf,martin2021nerf}, GANs employed 3D-aware generators to synthesize 3D representations~\cite{zhou2021cips,tian2018cr,gu2021stylenerf,chan2021pi}, which were further projected along query viewpoints into images. These methods were 
verified to be efficient to synthesize multi-view images with consistent 2D representations of 3D objects.
\section{Methodology}
In this paper, we propose a novel generative adversarial network (GAN) for \textbf{M}ulti-view \textbf{E}CG synthesis conditioned on heart diseases, called ME-GAN, which is composed of a two-stage generator and two discriminators (a major discriminator and a \textit{view discriminator}). Extending the idea of 3D-aware generators to ECG synthesis, our generator processes Gaussian noises $\mathbf{z}\in \mathbb{R}^d$ to synthesize the ``stereo'' electrocardio representations conditioned on heart diseases $\mathbf{c} \in \mathbb{R}^k$ (one of $k$ possible heart disease categories represented by a one-hot vector), and then employs the Nef-Net decoder \cite{chen2021electrocardio} to project the synthesized electrocardio representations into multi-view standard ECG signals. The major discriminator distinguishes real and synthesized ECG signals, and predicts the disease categories. The \textit{view discriminator} learns to revert randomly shuffled ECG views into a pre-determined order, which supervises the generator to synthesize ECG views representing proper view characteristics. Below we present the designs of these three key components of ME-GAN respectively.
\begin{figure*}[t]
\vskip -0.1 in
    \centering
    \includegraphics[width=0.95\textwidth]{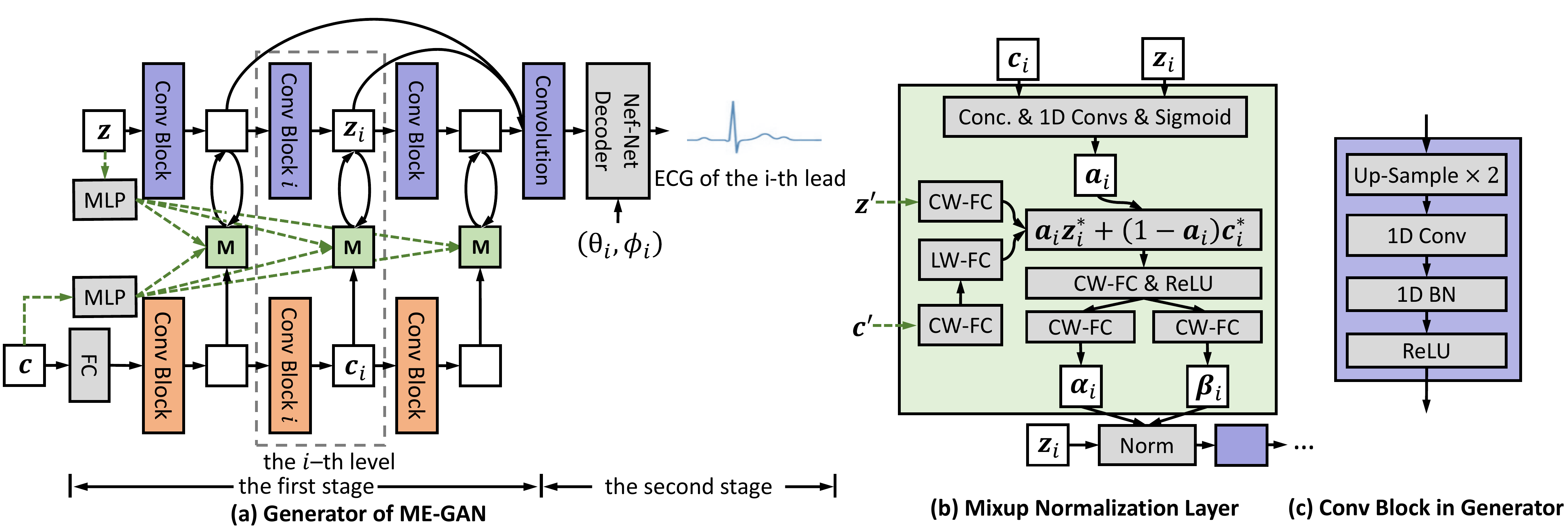}
    \vskip -0.15 in
    \caption{Illustrating the ME-GAN generator (a), in which ``M'' denotes a \textit{Mixup Normalization} layer as shown in (b) and the design of ``Conv Block'' is shown in (c). ``CW-FC'' and ``LW-FC'' denote a ``channel-wise fully connected layer'' and a ``length-wise fully connected layer'', respectively, and ``$\mathbf{z}$'' and ``$\mathbf{c}$'' represent Gaussian noises and disease condition, respectively. Notably, we build four levels in the first stage in our implementation.}
    \label{fig:framework}
    \vskip -0.15 in
\end{figure*}
\subsection{Generator}
\subsubsection*{The Overall Architecture} 
The generator architecture of ME-GAN has two stages (see Fig.~\ref{fig:framework}). The first stage synthesizes new ``stereo'' electrocardio representations using a ladder-shaped model; the second stage synthesizes multi-view ECG signals with a shallow 1D convolutional decoder of Nef-Net which takes the synthesized panoptic electrocardio representations and query viewpoint angles as input. Note that the input branch for deflection representation of the original Nef-Net decoder is not used here. The generator is performed end-to-end.

The ladder-shaped model of the first stage contains two model paths and some skip connections. The major path with 1D convolution blocks (marked in blue) processes and up-samples Gaussian noises $\mathbf{z} \in \mathbb{R}^d$ in steps, which obtains latent representations $\mathbf{z}_i \in \mathbb{R}^{d_i \times l_i}$ at the $i$-th level ($d_i$ and $l_i$ denote the feature channel number and feature length, respectively). All $\mathbf{z}_i$ were up-sampled to the final resolution and added together for the Nef-Net decoder. The other path (marked in orange) and two additional skip connections (marked by green dashed arrows) jointly yield the location-aware disease embedding and inject it into $\mathbf{z}_i$ via our proposed \textit{mixup normalization} layers (\textit{MixupNorm}), since the manifestations of heart diseases in ECG signals are localized in specific waveforms. Concretely, a disease condition $\mathbf{c} \in \mathbb{R}^{k}$ is first embedded into a latent code by a fully connected layer and is further processed and up-sampled by 1D convolution blocks into latent representations $\mathbf{c}_i \in \mathbb{R}^{k_i\times l_i}$ ($k_i$ indicates the channel size of $\mathbf{c}_i$). Apart from this, two multi-layer perceptrons (containing four fully connected layers with ReLU activations in between) respectively transform $\mathbf{z} \in \mathbb{R}^{d}$ and $\mathbf{c} \in \mathbb{R}^{k}$ into $\mathbf{z}^\prime \in \mathbb{R}^{d^\prime \times 1}$ and $\mathbf{c}^\prime \in \mathbb{R}^{k^\prime \times 1}$ ($d^\prime$ and $k^\prime$ indicate the channel sizes of $\mathbf{z}^\prime$ and $\mathbf{c}^\prime$, respectively), and feed them into \textit{MixupNorm} layers at all levels (see the green dashed arrows in Fig.~\ref{fig:framework}(a)). \textit{MixupNorm} uses $\mathbf{z}_i$ and $\mathbf{c}_i$ to compute a spatial attention to determine where and to what degree the disease condition (from $\mathbf{c}^\prime$) impacts the representation $\mathbf{z}_i$. As shown in Fig.~\ref{fig:framework}(c), a convolution block in the ladder-shaped model consists of a linear up-sample layer, a 1D convolution, a 1D batch normalization layer~\cite{ioffe2015batch}, and a ReLU activation. 
\subsubsection*{Mixup Normalization Operation} 
Without loss of generality, here we describe the operation of \textit{MixupNorm} at the $i$-th level (see Fig.~\ref{fig:framework}), since the operations at the other levels are similar. Similar to AdaIN~\cite{stylegan} and the self-modulation~\cite{chen2018self}, we inject condition information into the synthesized latent representation $\mathbf{z}_i$ by normalization operations.

\textit{MixupNorm} first concatenates the representation $\mathbf{z}_i \in \mathbb{R}^{d_i \times l_i}$ (which conveys the location information of the synthesized waveform) and $\mathbf{c}_i \in \mathbb{R}^{k_i \times l_i}$ (which conveys disease information) into a comprehensive representation of size $(d_i+k_i) \times l_i$, which is processed to attain an attention $\mathbf{a}_i \in \mathbb{R}^{d_a \times l_i}$ ($d_a$ is the channel size of $\mathbf{a}_i$). Intuitively, by fusing the information of the waveform location and disease information, the attention $\mathbf{a}_i$ can specify the degree to inject disease information at each location. Formally, $\mathbf{a}_i$ is computed by:
\begin{equation}\label{eq:attention}
    \mathbf{a}_i = \text{sigmoid}(\text{Conv}(\text{ReLU}(\text{BN}(\text{Conv}([\mathbf{z}_i, \mathbf{c}_i]))))),
\end{equation}
where ``$[\cdot, \cdot]$'' denotes channel-wise concatenation, and ``Conv'' and ``BN'' denote a 1D convolution layer and a 1D batch normalization layer, respectively. 
On the other hand, in \textit{MixupNorm}, the latent representation $\mathbf{z}^\prime$ is transformed by a channel-wise fully connected layer into $\mathbf{z}^*_i\in \mathbb{R}^{d_h\times 1}$, and $\mathbf{c}^\prime$ is transformed into $\mathbf{c}^*_i \in \mathbb{R}^{d_h \times l_i}$ by a channel-wise fully connected layer and a length-wise fully connected layer (with a ReLU in between), by:
\begin{equation}
    \mathbf{z}_i^* = {\mathbf{z}^\prime}^T U_i \ , \, \mathbf{c}_i^* = (\max(\mathbf{c}^\prime V_i + b_i, 0))^T W_i,
\end{equation}
where $U_i \in \mathbb{R}^{d^\prime \times d_h}$, $V_i,b_i \in \mathbb{R}^{1\times l_i}$, and $W_i \in \mathbb{R}^{k^\prime \times d_h}$ are learnable weights, and $d_h$ indicates the channel sizes of $\mathbf{z}^*_i$ and $\mathbf{c}^*_i$. Then, $\mathbf{z}_i^*$ is extended into size $(d_h\times l_i)$ and is incorporated with $\mathbf{c}_i^*$ according to attention $\mathbf{a}_i$, by:
\begin{equation}\label{eq:mixup}
  \mathbf{h}^*_i = \mathbf{a}_i \mathbf{c}_i^* + (\mathbf{1}-\mathbf{a}_i) \mathbf{z}_i^*,
\end{equation}
where $\mathbf{h}^*_i \in \mathbb{R}^{d_h\times l_i}$. A channel-wise fully connected layer then processes the representation $\mathbf{h}^*_i$ to synthesize the parameters $\alpha_i, \beta_i \in \mathbb{R}^{1 \times l_i}$ for normalization. Finally, $\alpha_i$ and $\beta_i$ are applied to the representation $\mathbf{z}_i$ in the major path, by:
\begin{equation}
    \widetilde{\mathbf{z}}_i = \alpha_i \odot \frac{\mathbf{z}_i - \mu}{\sigma} + \beta_i,
\end{equation}
where $\odot$ denotes point-wise multiplication, and $\mu\in \mathbb{R}^{1 \times l_i}$ and $\sigma\in \mathbb{R}^{1 \times l_i}$ are the mean and standard deviation of $\mathbf{z}_i$. The output $\widetilde{\mathbf{z}}_i$ goes forward to the next level. 

\textbf{Discussion.} In \textit{MixupNorm}, we use attention (in Eq.~(\ref{eq:attention})) to represent the locations and degrees for disease information injection. Note that $\mathbf{z}_i^* \in \mathbb{R}^{d_h\times 1}$ conveys the intrinsic style of noises similar to the self-modulation~\cite{chen2018self}, but $\mathbf{c}_i^*\in \mathbb{R}^{d_h\times l_i}$ represents different values along the length dimension. Unlike self-modulation, our \textit{MixupNorm} focuses on learning different disease embeddings at different locations in synthesized ECG signals explicitly.
\subsection{Discriminators}
\begin{figure}
\vskip -0.1 in
    \centering
    \includegraphics[width=0.5\textwidth]{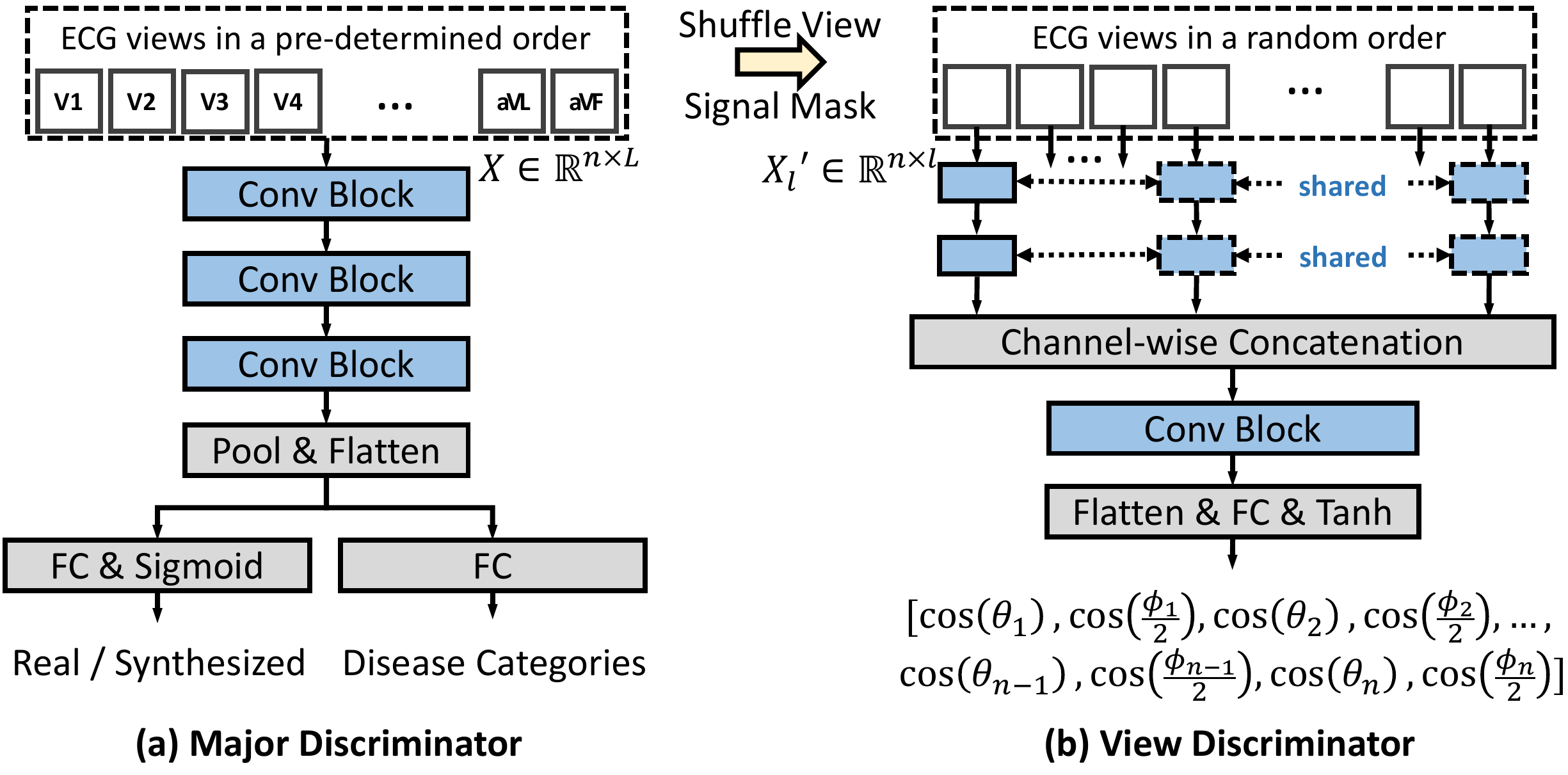}
    \vskip -0.15 in
    \caption{Illustrating the two ME-GAN discriminators. A blue rectangle denotes a convolution block consisting of a 1D convolution layer, an instance normalization, and a leaky ReLU. The input of the major discriminator is a tensor of size $n\times L$ containing all the $n$ ECG views, while the ECG signal input to the view discriminator is views shuffled and highly masked.}
    \vskip -0.1 in
    \label{fig:discriminators}
\end{figure}

Our ME-GAN employs two discriminators (see Fig.~\ref{fig:discriminators}(a)). The major discriminator takes ECG signals as input to determine whether they are real signals and their disease categories simultaneously, by employing LSGAN~\cite{lsgan} with an auxiliary classifier as in ACGAN~\cite{acgan}. The objective of the major discriminator $D$ and its supervision for generator $G$ are defined by:
\begin{equation}\label{eq:object1}
\begin{aligned}
        & \min_D \max_G \mathcal{V} (D, G) \\
        = & \mathbb{E}_{X\sim P_\text{data}} [(D_{\text{r/f}}(X)-a)^2 + \text{CE}(D_{c}(X), \mathbf{c})] \\
        & + \mathbb{E}_{X\sim P_\text{data};\mathbf{z}\sim P(\mathbf{z})}[(D_{\text{r/f}}(G(\mathbf{z}|\mathbf{c}))-b)^2 \\
        & + \text{BCE}(D^{\text{su}}_{c}(G(\mathbf{z}|\mathbf{c})), \mathbf{c})],
\end{aligned}
\end{equation}
where we set $a=1$, $b=0$ as default~\cite{lsgan}. $D_{\text{r/f}}$ indicates the major discriminator with the classifier head to predict whether the input is real, while $D_c$ indicates the major discriminator with the auxiliary classifier for disease classification (the superscript ``su'' denotes ``stop updation''). In Eq.~(\ref{eq:object1}), ``BCE'' denotes the binary cross-entropy loss function (since the heart disease classification is often a multi-class classification task), the tensor $X \in \mathbb{R}^{n \times L}$ contains ECG signals of length $L$ in all the $n$ views sampled from a dataset $P_\text{data}$, and $\mathbf{z}$ is sampled from Gaussian distribution $P_\mathbf{z}(\mathbf{z})$. The major discriminator is built with three convolution blocks (see Fig.~\ref{fig:discriminators}), and each block is built with a 1D convolution layer, a 1D instance normalization layer, and a leaky ReLU.
\subsubsection*{View Discriminator}
Unlike the major discriminator that takes all the ECG views organized in a pre-determined order as input, the input of the auxiliary discriminator (\textit{view discriminator}) contains all the ECG views given in a random order. In training, we randomly sample a permutation order of the views, and shuffle the ECG views and the corresponding target viewpoints (angles) according to the random order by the function $s_P$:
\begin{equation}\label{eq:shuffle}
    X^\prime = s_P(X) = (X^T P)^T, \, \Theta^\prime = s_P(\Theta) = (\Theta^T P)^T,
\end{equation}
where the $n$ views in $X \in \mathbb{R}^{n \times L}$ are organized in a pre-determined order. The cosines of the corresponding viewpoints are loaded in $\Theta \in \mathbb{R}^{n\times 2}$ by $\Theta[p, 1] = \text{cos}(\theta_p) \in (-1, 1]$ and $\Theta[p, 2] = \text{cos}(\phi_p/2) \in (-1, 1]$ ($p=1,2,\ldots, n$, the polar angle $\theta \in [0, \pi)$, and the azimuthal angle $\phi \in [0, 2\pi)$). The transition matrix $P$ of size $n\times n$ is a random doubly-stochastic matrix whose element values are 0 or 1. By multiplication with a random $P$, the orders of views and viewpoints in $X$ and $\Theta$ are shuffled into $X^\prime$ and $\Theta^\prime$. In our setting, the task of the \textit{view discriminator} is to take $X^\prime$ as input and predict the corresponding viewpoints $\Theta^\prime$. Since $P$ is random in each training epoch, correct viewpoint order predictions on synthesized ECG signals indicate that the synthesized ECG signals follow proper characteristics of the views. 
The \textit{view discriminator} employs a sub-network to iteratively process ECG signals in each view and concatenate the obtained representations to predict $2n$ values ranged from -1 to 1 (via a ``tanh'' operation on top of the model) to estimate the $2n$ values in $\Theta^\prime$. Instead of predicting the permutation matrix $P$ directly, we predict the angle information to avoid iterative processes in attaining the doubly-stochastic matrix $P$.

An obvious training strategy for the \textit{view discriminator} is to take $X^\prime$ as input and predict $\Theta^\prime$ directly. If the \textit{view discriminator} performs well on both real and synthesize ECG cases and obtains similar performances, then it means that the synthesized ECG views contain similar view characteristics as real ECG views. Yet, we need to further consider two issues. (1) The classification strategy can only verify that the view characteristics exist in the ECG views, but cannot prove that all the locations of the ECG signals represent correct view characteristics. (2) View characteristics in different locations vary. To address these two issues, we only feed a piece of ECG signals to the \textit{view discriminator}, and compare the predictions on the synthesized ECG signals and real ECG signals obtained by the \textit{view discriminator}. Formally, a piece of signals of a fixed length $l$ ($l < L$, where $L$ is the length of the original ECG signals) is selected for the \textit{view discriminator} by the function $t_\epsilon$, as:
\begin{equation}
    X^\prime_l = t_\epsilon(X^\prime) =X^\prime[:, \epsilon: \epsilon + l],
\end{equation}
where $\epsilon$ is randomly selected (subject to $\epsilon+l<L$) in each epoch, and $X^\prime_l \in \mathbb{R}^{n\times l}$. In addition, we conduct positional encoding (following the design in~\cite{vaswani2017attention}) on $X^\prime$ before it is processed by $t_\epsilon$. In training the \textit{view discriminator}, the \textit{view discriminator} processes real ECG signals $X^\prime_l$ to predict $\Theta^\prime$ under the guidance of mean square error (MSE). Meanwhile, it is required to enlarge the predictions of real and synthesized ECG signals, in order to push the \textit{view discriminator} to seek for some ``error view characteristics'' in the synthesized ECG views. In training the generator, the generator learns to reduce the prediction gap between real and synthesized ECG signals.
The objective function for the \textit{view discriminator} $V$ and its supervision for the generator $G$ are defined by:
\begin{equation}
\begin{aligned}
        & \min_V \max_G \mathcal{V}(V, G)\\
        =& \mathbb{E}_{X\sim P_\text{data}; \mathbf{z}\sim P_{\mathbf{z}}(\mathbf{z})}[\text{MSE}(V(t_\epsilon(s_P(X))), \Theta^\prime)\\ 
        & - \text{MSE}(V(t_\epsilon(s_P(X))|\mathbf{c}), V(t_\epsilon(s_P(G(\mathbf{z}|\mathbf{c})))))].
\end{aligned}
\end{equation}
\section{Relative Fr\'echet Inception Distance}
To measure the synthesis quality of GANs for ECG, we propose a new metric called \textit{relative Fr\'echet Inception Distance (rFID)}, inspired by the \textit{Fr\'echet Inception Distance (FID)}~\cite{heusel2017gans}. The \textit{FID} metric is used for imaging GANs to compute the distribution distance between a set of real images and a set of synthesized images, whose distributions are typically estimated by a pre-trained 2D Inception~\cite{szegedy2016rethinking}. According to this, we need a pre-trained model to estimate the distribution of multi-view ECG signals. A direct solution for this is to train a 1D Inception on a large-scale ECG dataset. However, we found that the open source ECG datasets are not enough to train a strongly generalized model in terms of the amount of data and the number of disease categories. For example, the MIT-BIH arrhythmia dataset has only 4 categories and 2 ECG views, while the Tianchi ECG dataset\footnote{\url{https://tianchi.aliyun.com/competition/entrance/231754/introduction}} has 34 categories.

Standard ECG data contain 12 views, and a length of 512 can well represent the waveform details. In this work, we set $n=12$ and $L=512$ ($X \in \mathbb{R}^{12\times 512}$) as default, and use a 1D Inception pre-trained on ImageNet~\cite{Deng2009}. Using the idea of Vision Transformer~\cite{dosovitskiy2020image}, we resize images into $46 \times 46$ and divide each image into 529 patches of size $2\times2 \times3$ (3 is the number of color channels). Thus, an image can be reorganized into size $12\times529$. Then, we drop the last 17 patches and obtain $12\times 512$.
In building the 1D Inception, we change 2D operations of the 2D Inception into the corresponding 1D operations. For a 2D convolution layer, we drop the last dimension of the kernel size (e.g., replace a $7\times1$ kernel by a 1D kernel of size 7). After pre-training the 1D Inception on the images, we use it to process ECG signals and obtain a latent representation of length 2048 (similar to the design of 2D Inception). To eliminate the incompatibility caused by using a model pre-trained on images to access the distribution of ECG data, we define a relative version of \textit{FID} (\textit{rFID}) by:
\begin{equation}\label{eq:rfid}
    \textit{rFID}(\{G(\mathbf{z})\}) = \frac{\textit{FID}(\{G(\mathbf{z}|\mathbf{c})|\mathbf{c} \in C_X\},\{X\}_2)}{\textit{FID}(\{X\}_1,\{X\}_2)},
\end{equation}
where the samples in a test set are divided into two groups $\{X\}_1$ and $\{X\}_2$ with identical distributions (e.g., the data amount of each category is identical), and $G$ is a trained GAN generator that synthesizes multi-view ECG signals with $\mathbf{z}\sim P_\mathbf{z}(\mathbf{z})$ and query conditions in $C_X$ that is identical to the condition distributions of $\{X\}_1$ and $\{X\}_2$. The denominator $\textit{FID}(\{X\}_1,\{X\}_2)$ is the \textit{FID} score between $\{X\}_1$ and $\{X\}_2$, which is used to reduce the gap between images and ECG. The effect of \textit{rFID} is verified in Sec.~\ref{exp:rfid}.
\section{Experiments}
\subsection{Experimental Setup}
\textbf{Datasets.}\ We conduct experiments on the Tianchi ECG dataset and PTB dataset~\cite{ptb} 
to evaluate our proposed ME-GAN. The Tianchi dataset contains 31,779 12-view ECG signals recorded at a frequency of 500 Hertz. The PTB dataset contains 549 12-view ECG signals recorded at a frequency of 1000 Hertz. On each dataset, we randomly take 80\% of ECG signals for training and the rest 20\% for test. 
We partition an ECG record into several cardiac cycles (representing heartbeats) using the partition annotations provided in~\cite{chen2021electrocardio}. Since an official heart disease annotation is for each ECG record containing several cardiac cycles, we use only some heart diseases that theoretically are present on all the cardiac cycles if occurred. Thus, for the Tianchi ECG dataset, we test our ME-GAN on 3 heart disease categories in two configurations: (1) with three disease conditions: the \textit{left axis deviation (LAD)}, the \textit{right axis deviation (RAD)}, and the \textit{right bundle branch block (\textit{RBBB})}; (2) without considering diseases. For the PTB dataset, we only test without using the heart disease conditions. Note that we use all the three selected diseases as conditions simultaneously in the tests on the Tianchi ECG dataset. For the configuration (1), we also use all the training data in the training phase, and the ECG data out of the three disease conditions were synthesized with input condition as a vector with zeros. All the cardiac cycles are used as independent samples, which are de-noised with the Python package~\cite{kachuee2018ecg}, interpolated to 500 Hertz, padded to length of $512$, and are linearly scaled to 0--1 in the pre-processing. We follow the viewpoint angle definitions given in~\cite{chen2021electrocardio} (12 views): ($\theta$, $\phi$) $\in \{$($\frac{\pi}{2}, \frac{\pi}{2}$), ($\frac{5\pi}{6}, \frac{\pi}{2}$), ($\frac{5\pi}{6}, -\frac{\pi}{2}$), ($\frac{\pi}{3}, -\frac{\pi}{2}$), ($\frac{\pi}{3}, \frac{\pi}{2}$), ($\pi, \frac{\pi}{2}$), ($\frac{\pi}{2}, -\frac{\pi}{18}$), ($\frac{\pi}{2}, \frac{\pi}{18}$), ($\frac{19\pi}{36}, \frac{\pi}{12}$), ($\frac{11\pi}{20}, \frac{\pi}{6}$), ($\frac{8\pi}{15}, \frac{\pi}{3}$), ($\frac{8\pi}{15}, \frac{\pi}{2}$)$\}$.
\begin{table*}[t]
\vskip -0.1 in
\setlength{\tabcolsep}{2pt}
\caption{\textbf{Synthesis performances of various GAN models}. The lower \textit{rFID} score is the better, and the accuracy score of 1-NN classifier (1NNC) is better if it is closer to 0.5. The best performances are marked in \textbf{bold}.}
\centering
\label{tab:synthesisperformances}
\begin{tabular}{l|cc|cc|cc|cc|cc|cc}
\toprule
\multicolumn{1}{c|}{\multirow{3}{*}{Methods}} & \multicolumn{8}{c|}{Tianchi w/ diseases}  &   \multicolumn{2}{c|}{Tianchi w/o diseases} &   \multicolumn{2}{c}{PTB w/o diseases} \\ \cmidrule{2-13}
\multicolumn{1}{c|}{}    &      \multicolumn{2}{c|}{overall}  &  \multicolumn{2}{c|}{\textit{LAD}}   & \multicolumn{2}{c|}{\textit{RBBB}} &  \multicolumn{2}{c|}{\textit{RAD}} & \multicolumn{2}{c|}{overall}&  \multicolumn{2}{c}{overall}\\
\cmidrule{2-13}
\multicolumn{1}{c|}{}      & \multicolumn{1}{c}{1NNC} &  \multicolumn{1}{c|}{\textit{rFID}} &  \multicolumn{1}{c}{1NNC} &  \multicolumn{1}{c|}{\textit{rFID}} &  \multicolumn{1}{c}{1NNC} &  \multicolumn{1}{c|}{\textit{rFID}} &  \multicolumn{1}{c}{1NNC} & \multicolumn{1}{c|}{\textit{rFID}} &  \multicolumn{1}{c}{1NNC} & \multicolumn{1}{c|}{\textit{rFID}} &  \multicolumn{1}{c}{1NNC} &   \multicolumn{1}{c}{\textit{rFID}}   \\ \midrule
WGAN-GP                  & 0.722 &  7.534  & 0.697 &  5.359  & 0.683 &  3.509   & 0.676 &   6.892  & 0.640 &  5.494 & 0.656 & 42.046 \\
ACGAN                    & 0.699 &  15.944  &  0.668 & 7.464 &  0.672 & 10.942  &  0.684 &  13.144 & -- & -- & -- & --\\
LSGAN    &  0.870 & 13.427  & 0.795 & 6.151 &  0.676 &  8.984   & 0.746 &  11.367   & 0.775 & 17.341  & 0.673 & 43.481  \\
CGAN        &  0.757 & 12.479   &  0.585 &  8.495 &  \textbf{0.607} & 5.489  &  \textbf{0.594} &  10.849 & -- & -- & -- & --  \\
SMDCGAN        & 0.998 & 38.619	& 0.934 &  23.985	 &  0.823 & 18.370   &  0.924 & 31.408   & 0.617 & 2.618  & 0.623 & 20.182 \\ \midrule
BC-GAN      & 0.832 & 12.701  & 0.723 & 8.299  & 0.617  & 5.975  & 0.634 &  10.685  & 0.983 &  42.545  & 0.997 & 158.698 \\
CBL-GAN   & 0.826  & 6.990 & 0.673  &   5.322  & 0.739  &   3.471  & 0.694 & 5.573  & 0.611 &   6.173  & 0.713  &  65.750      \\ \midrule
ME-GAN (Ours) & \textbf{0.643}  &   \textbf{3.722} & \textbf{0.567}  &  \textbf{2.662}  & 0.663    & \textbf{1.491} & 0.712   &     \textbf{3.433}  & \textbf{0.582}  &  \textbf{2.343} & \textbf{0.618}   &    \textbf{15.282} \\ \bottomrule
\end{tabular}
\end{table*}
\begin{table}[t]
\vskip -0.05 in
\centering
\setlength{\tabcolsep}{2pt}
\caption{\textbf{Classification performances of 1D ResNet-34 trained on augmented training sets.}}
\label{tab:classification}
\begin{tabular}{l|cccc}
\toprule
\multicolumn{1}{c|}{\multirow{2}{*}{Method}} & \multicolumn{4}{c}{Diseases (PR-AUC)} \\ \cmidrule{2-5} 
\multicolumn{1}{c|}{}                         & \textit{LAD}     & \textit{RBBB}     & \textit{RAD}  & Mean   \\ \midrule
\multicolumn{1}{c|}{1D ResNet-34 (baseline)}   &   0.925 &    0.801    &   0.911 &  0.879  \\
WGAN-GP      &  0.916   &   0.826    &   \textbf{0.922} & 0.889 \\
ACGAN    &    0.912     &     0.817     &    0.913  &   0.881     \\
LSGAN     &   0.918      &    0.858  &     0.910  & 0.895   \\
CGAN      &     0.906    &    0.867   &    0.904  &    0.892  \\
SMDCGAN          &    0.913     &    0.832      &    0.921  &    0.889  \\ \midrule
BC-GAN        &    0.896     &    0.820      &   0.898
 &  0.871    \\
CBL-GAN    &    0.910     &      0.857    &   0.896  &  0.888    \\ \midrule
ME-GAN (Ours)  &      \textbf{0.927}   &    \textbf{0.870}      &   0.908 &  \textbf{0.902} \\ \bottomrule
\end{tabular}
\vskip -0.25 in
\end{table}

\textbf{Implementation.}\ Our method is implemented with PyTorch 1.7 on an RTX2080Ti GPU. In training, the batch size is 16. For the generator, major discriminator, and \textit{view discriminator}, we use the learning rate $10^{-4}$ and the Adam optimizer with $\beta_1=0.5$, $\beta_2=0.999$. In the test without disease conditions, we replace the condition $c$ by a constant vector, and drop the auxiliary classifier in the major discriminator.

\textbf{Comparison Baselines.} We compare our ME-GAN with common GAN methods including conditional GAN (CGAN)~\cite{cgan}, WGAN-GP~\cite{gulrajani2017improved}, ACGAN~\cite{acgan}, LSGAN~\cite{lsgan}, and self-modulation DCGAN (SMDCGAN) with hinge loss~\cite{chen2018self}. Besides, we also compare the performances with state-of-the-art ECG synthesis methods, including BiLSTM-CNN GAN (BC-GAN)~\cite{zhu2019electrocardiogram} and CNN-BiLSTM-GAN (CBL-GAN)~\cite{delaney2019synthesis}. We use the GAN implementations in the projects\footnote{\url{https://github.com/znxlwm/pytorch-generative-model-collections}}$^,$\footnote{\url{https://github.com/MikhailMurashov/ecgGAN}} with few modifications to fit the data formats and to directly synthesize 12-view ECG signals. For the models that do not specify how conditions are used, we utilize the same strategy as in conditional GAN~\cite{cgan}. For the experimental settings without disease conditions, the input disease conditions and the related classifier (if a model has one) are not used. All the methods are trained in 100,000 iterations.
\begin{table*}[t]
\vskip -.15 in
\centering
\setlength{\tabcolsep}{2pt}
\caption{\textbf{Pairwise distance comparison of view signals within the same ECG cases.} For all the methods, 1000 samples were synthesized for the average pairwise distance computing.}
\label{tab:multiview}
\begin{tabular}{c|c|cccccccc}
\toprule
Methods            & Real ECG (baseline) & WGAN-GP & AC-GAN & LSGAN & CGAN  & SMDCGAN & BC-GAN & CBL-GAN & ME-GAN \\ \midrule
Distances & 0.259    & 0.327 & 0.482  & 0.435 & 0.423 & 0.843   & 0.405   & 0.355   & \textbf{0.286}  \\ 
\bottomrule
\end{tabular}
\vskip -.1 in
\end{table*}
\begin{table*}[t]
\vskip -0.1 in
\centering
\caption{\textbf{Ablation studies for the key components of our proposed ME-GAN.}}
\label{tab:ablationstudy}
\begin{tabular}{ccccc|cc}
\toprule
& \textit{View Discriminator} & Condition Injection & Auxiliary Classifier & Two-Stage Generator &  PR-AUC     & \textit{rFID}   \\ \midrule
(1)       & \checkmark   & \textit{MixNorm}       & \checkmark   & \checkmark & 0.902 & 3.722 \\
(2)     & \checkmark   & Self-modulation      & \checkmark  & \checkmark  &  0.858 & 5.432 \\
(3)               &    & \textit{MixNorm}      & \checkmark   & \checkmark   & 0.867 & 5.212   \\
(4)               & \checkmark      & \textit{MixNorm}       &          & \checkmark   &  0.895  &  4.580 \\ 
(5)    &  \checkmark     &   \textit{MixNorm}    &   \checkmark   &         &  0.884  &  4.585 \\\bottomrule
\end{tabular}
\vskip -0.2 in
\end{table*}
\subsection{Synthesis Performances}
To verify the quality of synthesized ECG signals, we report the \textit{rFID} scores and 1-NN classifier accuracy scores (1NNC) of various GAN methods in Table~\ref{tab:synthesisperformances}. One can see that the \textit{rFID} scores of our ME-GAN are much lower than those of the other methods on the two datasets, no matter with or without disease conditions. Specifically, comparing the performances of SMDCGAN and our method, ME-GAN outperforms SMDCGAN just a little if the disease conditions are not used (on both the Tianchi and PTB datasets). But, our method outperforms SMDCGAN with a clear margin if ECG syntheses are conditioned on the heart diseases. This might be because our proposed \textit{MixNorm} can better deal with the heart disease conditions than the self-modulation. We can also obtain the similar conclusion on 1-NN classifier accuracy scores.

Further, we evaluate the synthesis quality with classification performances (based on PR-AUC, i.e., the area under the precision-recall curve). We train 1D ResNet-34 models on the training sets augmented by the synthesized ECG signals, respectively. In training, we add synthesized samples to the original training set to double the size of the training data belonging to those three disease categories. The classification performances are reported in Table~\ref{tab:classification}. One can see that, compared to the performances obtained by an 1D ResNet-34 trained on the original training set (see the third row in Table~\ref{tab:classification}, mean PR-AUC $=$ 0.879), synthesized ECG signals promote the classification performances except for BiLSTM-CNN GAN. Among them, the synthesized ECG signals provided by our ME-GAN promote the classification performances by the most clear margins.

\subsection{Multi-view Consistency Validation}
To examine the multi-view consistency of synthesized ECG views, we compute and compare the pairwise distances of signals within an ECG case. We first train an 1D ResNet-34 (without the final fully connection layer) on the Tianchi ECG dataset with disease conditions supervised by the triplet loss~\cite{schroff2015facenet}, which independently processes each ECG signal of a view to attain an embedding, and learns to shorten the pairwise distances of the embedding from the same ECG case while to enlarge those distances from different cases. After training, the 1D ResNet is utilized to process each ECG signal on a view independently and obtains the corresponding embedding. We compute and report the average pairwise distances within ECG cases as in Table~\ref{tab:multiview}. The average pairwise distance computed on 1000 randomly selected cases from test sets is 0.259, while our ME-GAN obtains 0.286 which is pretty close to real ECG data (0.259). The performances of other methods were worse than ME-GAN, suggesting that ME-GAN can synthesize ECG signals whose waveforms on individual views were more consistent.
\subsection{Ablation Studies}
We conduct ablation studies to examine the effects of the key components of our ME-GAN on the Tianchi ECG dataset (with disease conditions), including the \textit{MixNorm}, \textit{view discriminator}, the auxiliary classifier used in the major discriminator, and the two-stage generator (with Nef-Net decoder). We evaluate \textit{MixNorm} by replacing it with the self-modulation~\cite{chen2018self}, and evaluate the two-stage generator by removing the Nef-Net decoder and using only the first stage part to synthesize 12 ECG views directly. The performances (based on \textit{rFID} and PR-AUC) are reported in Table~\ref{tab:ablationstudy}, and row (1) is ME-GAN with all the three components. It is evident that all the components provide performance gains for ME-GAN, and the \textit{view discriminator}, two-stage generator design, and \textit{MixNorm} are critically important. Besides, one can see that the performance ranks according to the mean PR-AUC and \textit{rFID} are the same, which suggests that the proposed \textit{rFID} is reliable.
\begin{figure*}[t]
    \centering
    \includegraphics[width=1\textwidth]{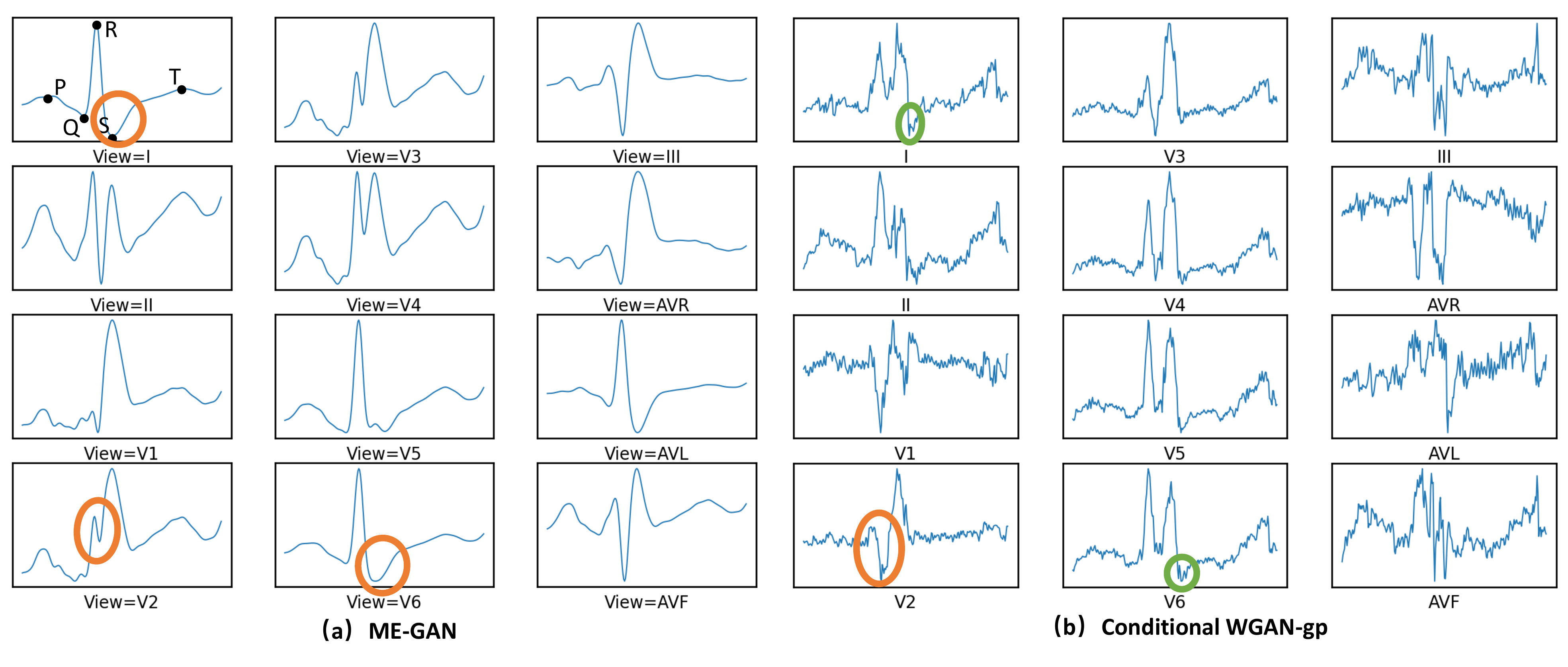}
    \vskip -0.25 in
    \caption{\textbf{Visualization comparison of synthesized 12-view ECG signals by ME-GAN and conditional WGAN-GP, conditioned on the \textit{RBBB} disease}. Orange circles mark some correct morbid manifestations of \textit{RBBB}, while green circles mark error manifestations.}
    \label{fig:vis}
\end{figure*}
\begin{figure*}[t]
\vskip -0.1 in
    \centering
    \includegraphics[width=\textwidth]{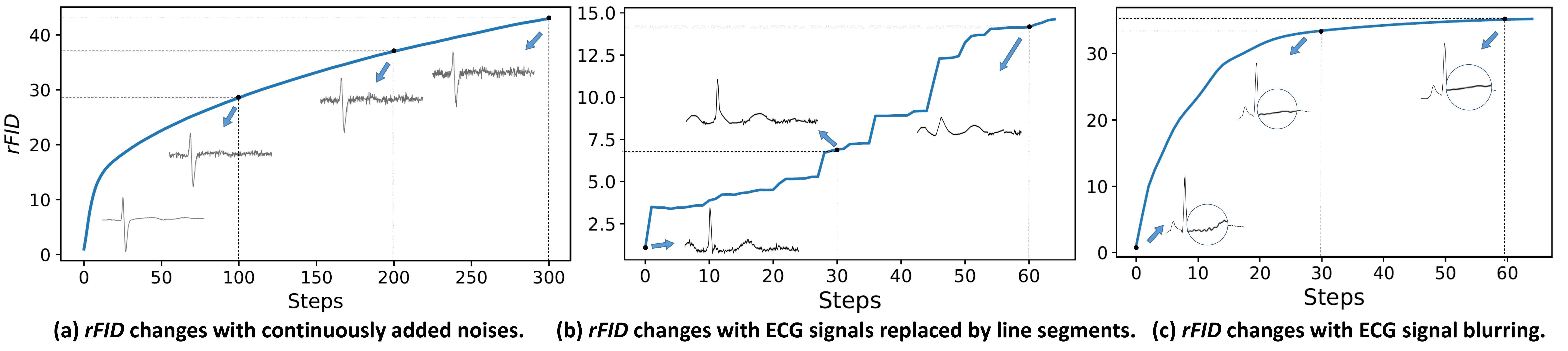}
    \vskip -0.2 in
    \caption{\textbf{Illustrating \textit{rFID} changes during a process in which noise is continually added onto ECG signals.} The ``V2'' view of an ECG signal is presented for visually showing the signal quality.}
    \label{fig:fid}
    \vskip -0.2 in
\end{figure*}
\subsection{Multi-view ECG Synthesis Visualization}
To intuitively show the synthesized ECG quality of ME-GAN, we visualize two cases of 12-view ECG signals conditioned on \textit{RBBB}, synthesized by ME-GAN and WGAN-GP for comparison (see Fig.~\ref{fig:vis}). It can be seen that our ME-GAN can synthesize smooth ECG signals, which represent clear waveforms (e.g., P, Q, R, S, T waves, as labeled by ``view$=$I'' in Fig.~\ref{fig:vis}(a)). In contrast, the conditional WGAN-GP does not synthesize clear waveforms, with noisy signals. Similar performances are seen by the other methods. There are two key manifestations of \textit{RBBB} in ECG signals: (1) a notch around R wave in the ``V2'' view; (2) the S waves in the ``V6'' and ``I'' views are of great duration, representing ``obtuse'' wave shapes (see Fig.~\ref{fig:vis}(a)). The synthesized ECG signals by our ME-GAN represent the manifestations (marked by orange circles) of both (1) and (2). While the ECG signals in the ``V2'' view synthesized by WGAN-GP (in Fig.~\ref{fig:vis}(b)) represent a notch around R wave (satisfying (1)), the S waves in the ``V1'' and ``V6'' views are of short duration (representing ``narrow'' and ``small'' S waves), which do not satisfy (2). Besides, we find that the signals in different views synthesized by conditional WGAN-GP look similar (e.g., in the ``V4'', ``V5'', and ``V6'' views), but similar phenomena do not occur in the synthesized ECG signals by our ME-GAN. We think that this might be because our two-stage major generator can learn the panoptic electrocardio representations for better multi-view ECG signal synthesis.
\subsection{Effects of the Metric \textit{rFID}}\label{exp:rfid}
Since the computation of \textit{rFID} in Eq.~(\ref{eq:rfid}) needs to divide the test set into two parts ($\{X\}_1, \{X\}_2$) with identical distribution to calculate the denominator (i.e., $\textit{FID}(\{X\}_1,\{X\}_2)$), we repeat computing the denominator with 10 different random divisions. We find that the values thus obtained are stable, with a standard deviation of $0.843$.  The same computation on the PTB dataset yields a standard deviation of $0.212$. This suggests that the denominator of \textit{rFID} is stable for a given dataset, and hence it can contribute to the consistency evaluation of ECG syntheses. Further, we observe the changes of $\textit{rFID}(\{X_\text{train}\})$, when we use some normalized real ECG data to replace $G(z)$ in Eq.~(\ref{eq:rfid}) and continually add various data perturbation. We test \textit{rFID} with three perturbation settings: in each step, (a) we add noise $\varepsilon \sim U(0,0.0002)$ onto the real ECG signals (normalized into 0--1); (b) we randomly erase 30 sequential sampling points on the ECG signal curves, and connect the onset and end breakpoints with a line segment; (c) we blur the ECG signals with an average blurring filter with $2\times 2$ kernel. As shown in Fig.~\ref{fig:fid}, \textit{rFID} scores increase with the perturbations continually added on, which indicates that \textit{rFID} is sensitive in assessing the quality of ECG signals.
\section{Conclusions}
In this paper, we proposed a novel disease-aware GAN, ME-GAN, which learns to obtain new panoptic electrocardio representations for multi-view ECG synthesis, applying the idea of 3D-aware generator to ECG synthesis and 
ensuring representation consistency among synthesized views. To make the synthesized ECG signals trusty, we also developed a novel \textit{view discriminator} which pushes synthesized views to represent proper view characteristics. A novel \textit{Mixup Normalization} layer injects disease information into synthesized waveforms adaptively, which contributes to synthesizing correct morbid manifestations. Besides, to directly assess the quality of synthesized ECG signals, we presented a pre-trained 1D Inception with a new metric \textit{rFID}. Comprehensive experiments showed that our designs are effective in multi-view ECG synthesis conditioned on heart diseases.
\section{Acknowledgements}
This research was partially supported by National Key R\&D Program of China under grant No.~2019YFC0118802, National Natural Science Foundation of China under grants No.~62176231 and 62106218, Zhejiang public welfare technology research project under grant No.~LGF20F020013. D. Z. Chen’s research was supported in part by NSF Grant CCF-1617735.
\bibliography{ref}

\begin{thebibliography}{55}
\providecommand{\natexlab}[1]{#1}
\providecommand{\url}[1]{\texttt{#1}}
\expandafter\ifx\csname urlstyle\endcsname\relax
  \providecommand{\doi}[1]{doi: #1}\else
  \providecommand{\doi}{doi: \begingroup \urlstyle{rm}\Url}\fi

\bibitem[Abubakar et~al.(2015)Abubakar, Tillmann, and
  Banerjee]{abubakar2015global}
Abubakar, I., Tillmann, T., and Banerjee, A.
\newblock Global, regional, and national age-sex specific all-cause and
  cause-specific mortality for 240 causes of death, 1990-2013: A systematic
  analysis for the global burden of disease study 2013.
\newblock \emph{The Lancet}, 2015.

\bibitem[Agrawal et~al.(2021)Agrawal, Venkitachalam,
  et~al.]{agrawal2021directional}
Agrawal, S., Venkitachalam, S., et~al.
\newblock Directional {GAN}: A novel conditioning strategy for generative
  networks.
\newblock \emph{ArXiv Preprint ArXiv:2105.05712}, 2021.

\bibitem[Arjovsky et~al.(2017)Arjovsky, Chintala, and Bottou]{wgan}
Arjovsky, M., Chintala, S., and Bottou, L.
\newblock Wasserstein generative adversarial networks.
\newblock In \emph{ICML}, 2017.

\bibitem[Bian et~al.(2022)Bian, Chen, et~al.]{bian2022identifying}
Bian, Y., Chen, J., et~al.
\newblock Identifying electrocardiogram abnormalities using a
  handcrafted-rule-enhanced neural network.
\newblock \emph{TCBB}, 2022.

\bibitem[Bousseljot et~al.(1995)Bousseljot, Kreiseler, and Schnabel]{ptb}
Bousseljot, R., Kreiseler, D., and Schnabel, A.
\newblock Nutzung der ekg-signaldatenbank cardiodat der ptb {\"u}ber das
  internet.
\newblock \emph{Biomedical Engineering / Biomedizinische Technik}, 1995.

\bibitem[Case et~al.(1979)]{case1979sequential}
Case, R.~B. et~al.
\newblock A sequential angular lead presentation.
\newblock \emph{Journal of Electrocardiology}, 1979.

\bibitem[Chan et~al.(2021)Chan, Monteiro, et~al.]{chan2021pi}
Chan, E.~R., Monteiro, M., et~al.
\newblock {Pi-GAN}: Periodic implicit generative adversarial networks for
  {3D}-aware image synthesis.
\newblock In \emph{CVPR}, 2021.

\bibitem[Chen et~al.(2021)Chen, Zheng, et~al.]{chen2021electrocardio}
Chen, J., Zheng, X., et~al.
\newblock Electrocardio panorama: Synthesizing new {ECG} views with
  self-supervision.
\newblock In \emph{IJCAI}, 2021.

\bibitem[Chen et~al.(2019)Chen, Lucic, et~al.]{chen2018self}
Chen, T., Lucic, M., et~al.
\newblock On self modulation for generative adversarial networks.
\newblock In \emph{ICLR}, 2019.

\bibitem[Chen et~al.(2016)Chen, Duan, et~al.]{chen2016infogan}
Chen, X., Duan, Y., et~al.
\newblock {InfoGAN}: Interpretable representation learning by information
  maximizing generative adversarial nets.
\newblock In \emph{NeurIPS}, 2016.

\bibitem[De~Cao \& Kipf(2018)De~Cao and Kipf]{molgan}
De~Cao, N. and Kipf, T.
\newblock {MolGAN}: An implicit generative model for small molecular graphs.
\newblock \emph{ArXiv Preprint ArXiv:1805.11973}, 2018.

\bibitem[Delaney et~al.(2019)Delaney, Brophy, et~al.]{delaney2019synthesis}
Delaney, A.~M., Brophy, E., et~al.
\newblock Synthesis of realistic {ECG} using generative adversarial networks.
\newblock \emph{ArXiv Preprint ArXiv:1909.09150}, 2019.

\bibitem[Deng et~al.(2009)Deng, Dong, Socher, Li, Li, and Li]{Deng2009}
Deng, J., Dong, W., Socher, R., Li, L.-J., Li, K., and Li, F.-F.
\newblock {ImageNet}: A large-scale hierarchical image database.
\newblock In \emph{CVPR}, 2009.

\bibitem[Dosovitskiy et~al.(2020)Dosovitskiy, Beyer,
  et~al.]{dosovitskiy2020image}
Dosovitskiy, A., Beyer, L., et~al.
\newblock An image is worth 16x16 words: {Transformers} for image recognition
  at scale.
\newblock In \emph{ICLR}, 2020.

\bibitem[Frank(1956)]{franklead}
Frank, E.
\newblock An accurate, clinically practical system for spatial
  vectorcardiography.
\newblock \emph{Circulation}, 1956.

\bibitem[Golany \& Radinsky(2019)Golany and Radinsky]{golany2019pgans}
Golany, T. and Radinsky, K.
\newblock {PGANs}: Personalized generative adversarial networks for {ECG}
  synthesis to improve patient-specific deep {ECG} classification.
\newblock In \emph{AAAI}, 2019.

\bibitem[Golany et~al.(2020)Golany, Radinsky, and Freedman]{golany2020simgans}
Golany, T., Radinsky, K., and Freedman, D.
\newblock {SimGANs}: Simulator-based generative adversarial networks for {ECG}
  synthesis to improve deep {ECG} classification.
\newblock In \emph{ICML}, 2020.

\bibitem[Golany et~al.(2021)Golany, Radinsky, et~al.]{golany202112}
Golany, T., Radinsky, K., et~al.
\newblock 12-lead {ECG} reconstruction via {Koopman} operators.
\newblock In \emph{ICML}, 2021.

\bibitem[Goodfellow et~al.(2014)Goodfellow, Pouget-Abadie,
  et~al.]{goodfellow2014generative}
Goodfellow, I., Pouget-Abadie, J., et~al.
\newblock Generative adversarial nets.
\newblock \emph{NeurIPS}, 2014.

\bibitem[Grant(1950)]{grant1950spatial}
Grant, R.~P.
\newblock Spatial vector electrocardiography: A method for calculating the
  spatial electrical vectors of the heart from conventional leads.
\newblock \emph{Circulation}, 1950.

\bibitem[Graybiel et~al.(1946)Graybiel, White,
  et~al.]{graybiel1946electrocardiography}
Graybiel, A., White, P.~D., et~al.
\newblock \emph{Electrocardiography in practice}.
\newblock Saunders, 1946.

\bibitem[Gu et~al.(2021)Gu, Liu, et~al.]{gu2021stylenerf}
Gu, J., Liu, L., et~al.
\newblock {StyleNerf}: A style-based {3D}-aware generator for high-resolution
  image synthesis.
\newblock \emph{ArXiv Preprint ArXiv:2110.08985}, 2021.

\bibitem[Gulrajani et~al.(2017)Gulrajani, Ahmed, et~al.]{gulrajani2017improved}
Gulrajani, I., Ahmed, F., et~al.
\newblock Improved training of wasserstein {GANs}.
\newblock \emph{ArXiv Preprint ArXiv:1704.00028}, 2017.

\bibitem[Hazra \& Byun(2020)Hazra and Byun]{hazra2020synsiggan}
Hazra, D. and Byun, Y.-C.
\newblock {SynSigGAN}: Generative adversarial networks for synthetic biomedical
  signal generation.
\newblock \emph{Biology}, 2020.

\bibitem[Heusel et~al.(2017)Heusel, Ramsauer, et~al.]{heusel2017gans}
Heusel, M., Ramsauer, H., et~al.
\newblock {GANs} trained by a two time-scale update rule converge to a local
  {Nash} equilibrium.
\newblock \emph{NeurIPS}, 2017.

\bibitem[Holst et~al.(1999)Holst, Ohlsson, et~al.]{holst1999confident}
Holst, H., Ohlsson, M., et~al.
\newblock A confident decision support system for interpreting
  electrocardiograms.
\newblock \emph{Clinical Physiology}, 1999.

\bibitem[Hossain et~al.(2021)Hossain, Kamran, et~al.]{ecgadvgan}
Hossain, K.~F., Kamran, S.~A., et~al.
\newblock {ECG-Adv-GAN}: Detecting {ECG} adversarial examples with conditional
  generative adversarial networks.
\newblock \emph{ArXiv Preprint ArXiv:2107.07677}, 2021.

\bibitem[Huang \& Belongie(2017)Huang and Belongie]{huang2017arbitrary}
Huang, X. and Belongie, S.
\newblock Arbitrary style transfer in real-time with adaptive instance
  normalization.
\newblock In \emph{ICCV}, 2017.

\bibitem[Ioffe \& Szegedy(2015)Ioffe and Szegedy]{ioffe2015batch}
Ioffe, S. and Szegedy, C.
\newblock Batch normalization: Accelerating deep network training by reducing
  internal covariate shift.
\newblock In \emph{ICML}, 2015.

\bibitem[Kachuee et~al.(2018)Kachuee, Fazeli, and Sarrafzadeh]{kachuee2018ecg}
Kachuee, M., Fazeli, S., and Sarrafzadeh, M.
\newblock {ECG} heartbeat classification: A deep transferable representation.
\newblock In \emph{ICHI}, 2018.

\bibitem[Karras et~al.(2019)Karras, Laine, and Aila]{stylegan}
Karras, T., Laine, S., and Aila, T.
\newblock A style-based generator architecture for generative adversarial
  networks.
\newblock In \emph{CVPR}, 2019.

\bibitem[Karras et~al.(2020)Karras, Laine, et~al.]{stylegan2}
Karras, T., Laine, S., et~al.
\newblock Analyzing and improving the image quality of {StyleGAN}.
\newblock In \emph{CVPR}, 2020.

\bibitem[Kiranyaz et~al.(2015)Kiranyaz, Ince, and Gabbouj]{kiranyaz2015real}
Kiranyaz, S., Ince, T., and Gabbouj, M.
\newblock Real-time patient-specific {ECG} classification by 1-{D}
  convolutional neural networks.
\newblock \emph{IEEE Transactions on Biomedical Engineering}, 2015.

\bibitem[Kuznetsov et~al.(2020)Kuznetsov, Moskalenko, and
  Zolotykh]{kuznetsov2020electrocardiogram}
Kuznetsov, V., Moskalenko, V., and Zolotykh, N.~Y.
\newblock Electrocardiogram generation and feature extraction using a
  variational autoencoder.
\newblock \emph{ArXiv Preprint ArXiv:2002.00254}, 2020.

\bibitem[Kuznetsov et~al.(2021)Kuznetsov, Moskalenko,
  et~al.]{kuznetsov2021interpretable}
Kuznetsov, V., Moskalenko, V., et~al.
\newblock Interpretable feature generation in {ECG} using a variational
  autoencoder.
\newblock \emph{Frontiers in Genetics}, 2021.

\bibitem[Mao et~al.(2017)Mao, Li, et~al.]{lsgan}
Mao, X., Li, Q., et~al.
\newblock Least squares generative adversarial networks.
\newblock In \emph{ICCV}, 2017.

\bibitem[Martin-Brualla et~al.(2021)Martin-Brualla, Radwan,
  et~al.]{martin2021nerf}
Martin-Brualla, R., Radwan, N., et~al.
\newblock {NeRF} in the wild: Neural radiance fields for unconstrained photo
  collections.
\newblock In \emph{CVPR}, 2021.

\bibitem[McSharry et~al.(2003)McSharry, Clifford,
  et~al.]{mcsharry2003dynamical}
McSharry, P.~E., Clifford, G.~D., et~al.
\newblock A dynamical model for generating synthetic electrocardiogram signals.
\newblock \emph{IEEE Transactions on Biomedical Engineering}, 2003.

\bibitem[Mildenhall et~al.(2020)Mildenhall, Srinivasan,
  et~al.]{mildenhall2020nerf}
Mildenhall, B., Srinivasan, P.~P., et~al.
\newblock {NeRF}: Representing scenes as neural radiance fields for view
  synthesis.
\newblock In \emph{ECCV}, 2020.

\bibitem[Mirza \& Osindero(2014)Mirza and Osindero]{cgan}
Mirza, M. and Osindero, S.
\newblock Conditional generative adversarial nets.
\newblock \emph{ArXiv Preprint ArXiv:1411.1784}, 2014.

\bibitem[Odena(2016)]{odena2016semi}
Odena, A.
\newblock Semi-supervised learning with generative adversarial networks.
\newblock \emph{ArXiv Preprint ArXiv:1606.01583}, 2016.

\bibitem[Odena et~al.(2017)Odena, Olah, and Shlens]{acgan}
Odena, A., Olah, C., and Shlens, J.
\newblock Conditional image synthesis with auxiliary classifier {GANs}.
\newblock In \emph{ICML}, 2017.

\bibitem[Radford et~al.(2015)Radford, Metz, and
  Chintala]{radford2015unsupervised}
Radford, A., Metz, L., and Chintala, S.
\newblock Unsupervised representation learning with deep convolutional
  generative adversarial networks.
\newblock \emph{ArXiv Preprint ArXiv:1511.06434}, 2015.

\bibitem[Ramsundar et~al.(2015)Ramsundar, Kearnes,
  et~al.]{ramsundar2015massively}
Ramsundar, B., Kearnes, S., et~al.
\newblock Massively multitask networks for drug discovery.
\newblock \emph{ArXiv Preprint ArXiv:1502.02072}, 2015.

\bibitem[Roth et~al.(2018)Roth, Abate, et~al.]{roth2018global}
Roth, G.~A., Abate, D., et~al.
\newblock Global, regional, and national age-sex-specific mortality for 282
  causes of death in 195 countries and territories, 1980--2017: A systematic
  analysis for the global burden of disease study 2017.
\newblock \emph{The Lancet}, 2018.

\bibitem[Schroff et~al.(2015)Schroff, Kalenichenko, and
  Philbin]{schroff2015facenet}
Schroff, F., Kalenichenko, D., and Philbin, J.
\newblock {FaceNet}: {A} unified embedding for face recognition and clustering.
\newblock In \emph{CVPR}, 2015.

\bibitem[Sutskever et~al.(2014)Sutskever, Vinyals, and
  Le]{sutskever2014sequence}
Sutskever, I., Vinyals, O., and Le, Q.~V.
\newblock Sequence to sequence learning with neural networks.
\newblock In \emph{NeurIPS}, 2014.

\bibitem[Szegedy et~al.(2016)Szegedy, Vanhoucke, et~al.]{szegedy2016rethinking}
Szegedy, C., Vanhoucke, V., et~al.
\newblock Rethinking the {Inception} architecture for computer vision.
\newblock In \emph{CVPR}, 2016.

\bibitem[Thambawita et~al.(2021)Thambawita, Isaksen,
  et~al.]{thambawita2021deepfake}
Thambawita, V., Isaksen, J.~L., et~al.
\newblock {DeepFake} electrocardiograms using generative adversarial networks
  are the beginning of the end for privacy issues in medicine.
\newblock \emph{Scientific Reports}, 2021.

\bibitem[Tian et~al.(2018)Tian, Peng, et~al.]{tian2018cr}
Tian, Y., Peng, X., et~al.
\newblock {CR-GAN}: Learning complete representations for multi-view
  generation.
\newblock \emph{ArXiv Preprint ArXiv:1806.11191}, 2018.

\bibitem[Vaswani et~al.(2017)Vaswani, Shazeer, Parmar,
  et~al.]{vaswani2017attention}
Vaswani, A., Shazeer, N., Parmar, N., et~al.
\newblock Attention is all you need.
\newblock \emph{NeurIPS}, 2017.

\bibitem[Yi et~al.(2019)Yi, Walia, and Babyn]{yi2019generative}
Yi, X., Walia, E., and Babyn, P.
\newblock Generative adversarial network in medical imaging: A review.
\newblock \emph{Medical Image Analysis}, 2019.

\bibitem[Zhang \& Babaeizadeh(2021)Zhang and Babaeizadeh]{zhang2021synthesis}
Zhang, Y.-H. and Babaeizadeh, S.
\newblock Synthesis of standard 12-lead electrocardiograms using two
  dimensional generative adversarial network.
\newblock \emph{ArXiv Preprint ArXiv:2106.03701}, 2021.

\bibitem[Zhou et~al.(2021)Zhou, Xie, et~al.]{zhou2021cips}
Zhou, P., Xie, L., et~al.
\newblock {CIPS-3D: A 3D}-aware generator of {GANs} based on
  conditionally-independent pixel synthesis.
\newblock \emph{ArXiv Preprint ArXiv:2110.09788}, 2021.

\bibitem[Zhu et~al.(2019)Zhu, Ye, et~al.]{zhu2019electrocardiogram}
Zhu, F., Ye, F., et~al.
\newblock Electrocardiogram generation with a bidirectional {LSTM-CNN}
  generative adversarial network.
\newblock \emph{Scientific Reports}, 2019.

\end{thebibliography}
\bibliographystyle{icml2022}



\end{document}